\documentclass[10pt,twocolumn,letterpaper]{article}

\usepackage[pagenumbers]{cvpr} % To force page numbers, e.g. for an arXiv version

\usepackage{graphicx}
\usepackage{amsmath}
\usepackage{amssymb}
\usepackage{booktabs}
\usepackage{multirow}
\usepackage{subcaption}
\usepackage{enumitem}
\usepackage{array}
\usepackage{siunitx}
\newcolumntype{C}[1]{>{\centering\let\newline\\\arraybackslash\hspace{0pt}}p{#1}}
\usepackage[pagebackref,breaklinks,colorlinks]{hyperref}

\usepackage[capitalize]{cleveref}
\crefname{section}{Sec.}{Secs.}
\Crefname{section}{Section}{Sections}
\Crefname{table}{Table}{Tables}
\crefname{table}{Tab.}{Tabs.}

\def\cvprPaperID{1} % *** Enter the CVPR Paper ID here
\def\confName{OmniCV}
\def\confYear{2023}

\begin{document}

\title{Human Pose Estimation in Monocular Omnidirectional Top-View Images}

\author{Jingrui Yu\(^*\)\quad Tobias Scheck\quad Roman Seidel\quad
Yukti Adya\quad Dipankar Nandi\quad Gangolf Hirtz\\
Chemnitz University of Technology, Germany\\
\tt\small \(^*\)jingrui.yu@etit.tu-chemnitz.de
}
\maketitle

%%%%%%%%% ABSTRACT
%%%%%%%%% ABSTRACT
\begin{abstract}
   Human pose estimation (HPE) with convolutional neural networks (CNNs) for indoor monitoring is one of the major challenges in computer vision.
   In contrast to HPE in perspective views, an indoor monitoring system can consist of an omnidirectional camera with a field of view of 180° to detect the pose of a person with only one sensor per room.
   To recognize human pose, the detection of keypoints is an essential upstream step.
   In our work we propose a new dataset for training and evaluation of CNNs for the task of keypoint detection in omnidirectional images.
   The training dataset, THEODORE+, consists of 50,000 images and is created by a 3D rendering engine, where humans are randomly walking through an indoor environment. 
   In a dynamically created 3D scene, persons move randomly with simultaneously moving omnidirectional camera to generate synthetic RGB images and 2D and 3D ground truth. 
   For evaluation purposes, the real-world PoseFES dataset with two scenarios and 701 frames with up to eight persons per scene was captured and annotated. 
   We propose four training paradigms to finetune or re-train two top-down models in MMPose and two bottom-up models in CenterNet on THEODORE+.
   Beside a qualitative evaluation we report quantitative results.
   Compared to a COCO pretrained baseline, we achieve significant improvements especially for top-view scenes on the PoseFES dataset.
   Our datasets can be found at \url{https://www.tu-chemnitz.de/etit/dst/forschung/comp_vision/datasets/index.php.en}.
\end{abstract}

%%%%%%%%% BODY TEXT
\section{Introduction}
\label{sec:intro}
With the growing need for indoor monitoring, human pose estimation (HPE) has become one of the main research areas of modern computer vision research \cite{meinel2014automated,richter2015pose}.
However, most of this research was conducted on perspective views with a limited field of view which is not appropriate for indoor scenarios when an entire room is to be covered with only one sensor.
For this problem, cameras with an omnidirectional camera model and wide-angle fisheye lenses are best suited. 
Nevertheless, fish-eye lenses have the characteristic of distortion which makes it necessary to model these distortions implicitly in the training data.
In this work, we propose a fuss free way to use convolutional neural networks (CNNs) for human pose estimation in omnidirectional images.
% TODO: Forschungsfrage und Ziel der Arbeit nennen.
The application field of this work is Ambient Assisted Living (AAL).
AAL stands for concepts, products and services that introduce new technologies into daily living in order to improve the quality of life for people in all phases of life, especially in old age.
These systems, here a camera-based smart sensor that is mounted in the ceiling of a room, monitor elderlies' activities to provide behaviour analysis results for care attendants, relatives and physicians.
Besides a daily activity protocol, the system is capable to recognize emergency situations such as unpredictable human falls \cite{seidel2020contactless}.
With the help of infrared lighting, video-based sleep quality estimation is employed to be able to draw conclusions about the current behaviour of an affected person as well as anomalies of the behaviour.
The contribution of our work is threefold:
\begin{itemize}
 \item we provide a top-view omnidirectional synthetic dataset with keypoint annotations, namely THEODORE+.
 \item For evaluation purposes, the \textit{PoseFES} dataset, a new real-world top-view omnidirectional dataset with keypoint annotations was created.
 \item The adaption of state-of-the-art approaches for keypoint detection to a monocular top-view omnidirectional camera.
\end{itemize}
% The remainder of the paper is structured as follows: chapter 2 describes related works in the field of HPE and synthetic datasets especially for omnidirectional vision.
% Chapter 3 explains the dataset used in our work, both a synthetic and real-world fisheye dataset with keypoint annotations, namely THEODORE+ and PoseFES.
% In chapter 4 the model choices, training schedule and experiments of our work is described, where chapter 5 shows results on the real-world dataset PoseFES. 
% The paper closes with the conclusion in chapter 6.

\section{Related work}
\label{sec:sota}

\subsection{Human pose estimation in perspective and omnidirectional images}
HPE is a popular topic in computer vision because of its wide range of applications, such as action and activity recognition, augmented reality (AR) and rehabilitation feedback systems.
Early implementations have used hand-crafted features as well as pre-defined human models \cite{s16121966}.
In recent years, researchers have shifted to deep learning-based methods.
Xiao \etal \cite{xiao2018simple} provided a simple baseline by using ResNet-50 \cite{He_2016_CVPR} to generate heatmaps for each keypoint.
HRNet \cite{sun2019deep} further improved the results by introducing a high resolution backbone network.
These are top-down methods, which use a separate person detector to isolate image areas with the person, and then perform keypoint estimation on them.
In contrast, bottom-up methods are able to detect keypoints for multiple persons in a single inference pipeline.
OpenPose \cite{cao2017realtime} uses part affinity fields to associate the estimated keypoints with the individual persons in the image.
CenterNet \cite{zhou2019objects} estimates the offset of each joint to the person center point and use it to associate them with the persons.

Though HPE is well researched for perspective images, it is only starting to get attention for omnidirectional images.
There has been research that focus on using head-mounted fisheye cameras for pose estimation, such as \textit{EgoCap} \cite{rhodin2016egocap}, \textit{Mo\textsuperscript{2}Cap\textsuperscript{2}} \cite{xu2019mo}, \textit{xR-EgoPose} \cite{tome2019xr}\textit{\,/\,SelfPose} \cite{tome2020selfpose}, \textit{EgoGlass} \cite{zhao2021egoglass} and Wang \etal \cite{wang2021estimating}.
However, the application of CNNs for HPE in overhead omnidirectional images has not been thoroughly investigated.
This is largely due to the lack of training data in this domain.

Georgakopoulos \etal \cite{delibasis2016geodesically,georgakopoulos2018pose} employ a 3D human model to create a dataset of binary silhouettes, which are rendered through the calibration of the fisheye camera.
The CNN is trained to differentiate between the pre-set poses, rather than estimate the joint positions.
Haque \etal \cite{haque2016vpinvariant3dhpe} train CNN and LSTM \cite{hochreiter1997lstm}  to achieve view-point invariant 3d pose estimation on a singular depth image.
Denecke and Jauch \cite{denecke2021verification} use the 3D point cloud calculated by the smart sensor and prior knowledge of the human body to estimate the joint positions.
The results of this method are restricted by factors such as the mounting position of the camera and differences between each individual body.
The inference speed is limited by the speed of the smart sensor.
Heindl \etal \cite{heindl2019large} generate multiple rectilinear views from a fisheye image and perform 2D keypoint estimation using OpenPose without finetuning the model.
The 3D skeleton is reconstructed by a stereo vision setup.
This method requires that the calibration parameters of the cameras are known.
The temporal performance is heavily restricted by the overhead of generating multiple views and inferencing on all of them.
Garau \etal achieve viewpoint-invariant 3D HPE with a capsule auto-encoder named DECA \cite{garau2021deca} on depth and RGB images, namely ITOP and PanopTOP31K datasets.

\subsection{Top-view HPE datasets}
\label{subsec:tvdatasets}
Haque \etal \cite{haque2016vpinvariant3dhpe} introduced the Invariant-Top View Dataset (ITOP), which consists of 100K real-world depth images.
It contains no RGB-images, therefore limiting the use of popular CNN-based HPE models.
Garau \etal introduced the PanopTOP framework in \cite{garau2021panoptop} for generating semi-synthetic top-view human images of normal perspective camera with 2D- and 3D-pose groundtruth from the multi-view dataset Panoptic \cite{joo2015panoptic}.
With this framework they create the PanopTOP31K dataset.
It contains top-view and front-view persons as well as the corresponding depth maps, point clouds and 3D meshes.
This is the first semi realistic HPE dataset of RGB images from the top-view.
However, there are a few shortcomings in this dataset. 
Firstly, the image resolution is very low at \(256\times 256\) pixels, while the persons in the images effectively occupy no more than \(100\times100\) pixels.
Secondly, there are a lot of artifacts in the synthesized images, the most severe of which is hand position ghosting, where there are multiple instances of each arm\,/\,hand in the images.
Finally, the subjects are positioned in front of a white background and the camera position is fixed, thus the variations are low across the dataset.

\subsection{Synthetic data generation for CNN training}
There exist many datasets for the task HPE, such as MPII \cite{mpii2014}, Human3.6M \cite{h36m} and COCO keypoints \cite{lin2014microsoft}.
However, capturing real-world HPE data with accurate groundtruth annotations either require a specific motion capture system \cite{h36m} or a large-scale manual annotation process \cite{lin2014microsoft}.
Therefore, it is often expensive and time-consuming.
Synthetic data generation, on the other hand, can create pixel precise annotations without additional steps.
Additionally, it does not raise privacy concerns.

Song \etal \cite{songSemanticSceneCompletion2017} introduce 3D computer graphics to create interior scenes with realistic textures and furniture.
The authors in \cite{wuBuildingGeneralizableAgents2018} extended this concept with agent functionality to navigate freely in the 3D environment.
McCormac \etal \cite{mccormacScenenetRgbdCan2017} describe methods for creating physical and photorealistic interior renderings.
A similar approach is taken by Li \etal \cite{liInteriorNetMegascaleMultisensor2018}, but with a vast amount of professional interior designs and object assets while creating a synthetic dataset.
In \cite{chenSynthesizingTrainingImages2016} the authors simulate different human bodies with various assigned textures to train CNNs to recognize poses.
Hoffman \etal \cite{hoffmannLearningTrainSynthetic2019} compare synthetically generated humans to investigate the influence of synthetic data compared to real but augmented data.
With SURREAL \cite{varolLearningSyntheticHumans2017}, the authors introduce a dataset with photorealistic computer-generated images containing annotations for body parts and action sequences.
In a multi-agent simulation concerning interaction in household scenarios, Puig \etal \cite{puigVirtualhomeSimulatingHousehold2018} present VirtualHome.
Since VirtualHome provides interactions of humans with objects, this simulation serves very well in an AAL context.
With ElderSIM \cite{hwangElderSimSyntheticData2020}, another AAL-focused eldercare simulation is introduced.
Here various modalities, including RGB videos, 2D and 3D skeletons and different camera angles, are provided.
Concerning omnidirectional imagery, none of the above methods provide such content, which is mandatory if CNNs are to be applied in this domain.
A simulation using computer graphics for omnidirectional ego-pose imagery is introduced in \cite{rhodin2016egocap,tome2019xr,tome2020selfpose}.

To obtain omnidirectional top-view images to train CNNs, the work of Scheck \etal \cite{scheck2020learning} introduces a procedure for generating such images, resulting in a dataset for object detection and semantic segmentation.
This approach was extended by Seuffert \etal \cite{seuffert2021study} with stereo image pairs and the corresponding depth maps.
Seidel \etal \cite{seidel2021omniflow} present an omnidirectional dataset for the optical flow, focusing on household activities.
However, skeletal keypoints for simulated humans are still missing in these datasets for the application of HPE.

\section{Datasets}
\label{sec:datasets}

\subsection{THEODORE+ dataset}
\label{subsec:theo}

\begin{figure*}
    \centering
    \subcaptionbox{RGB Image \label{subfig:theodore_plus:rgb}}[.32\linewidth]{\includegraphics[width=\linewidth]{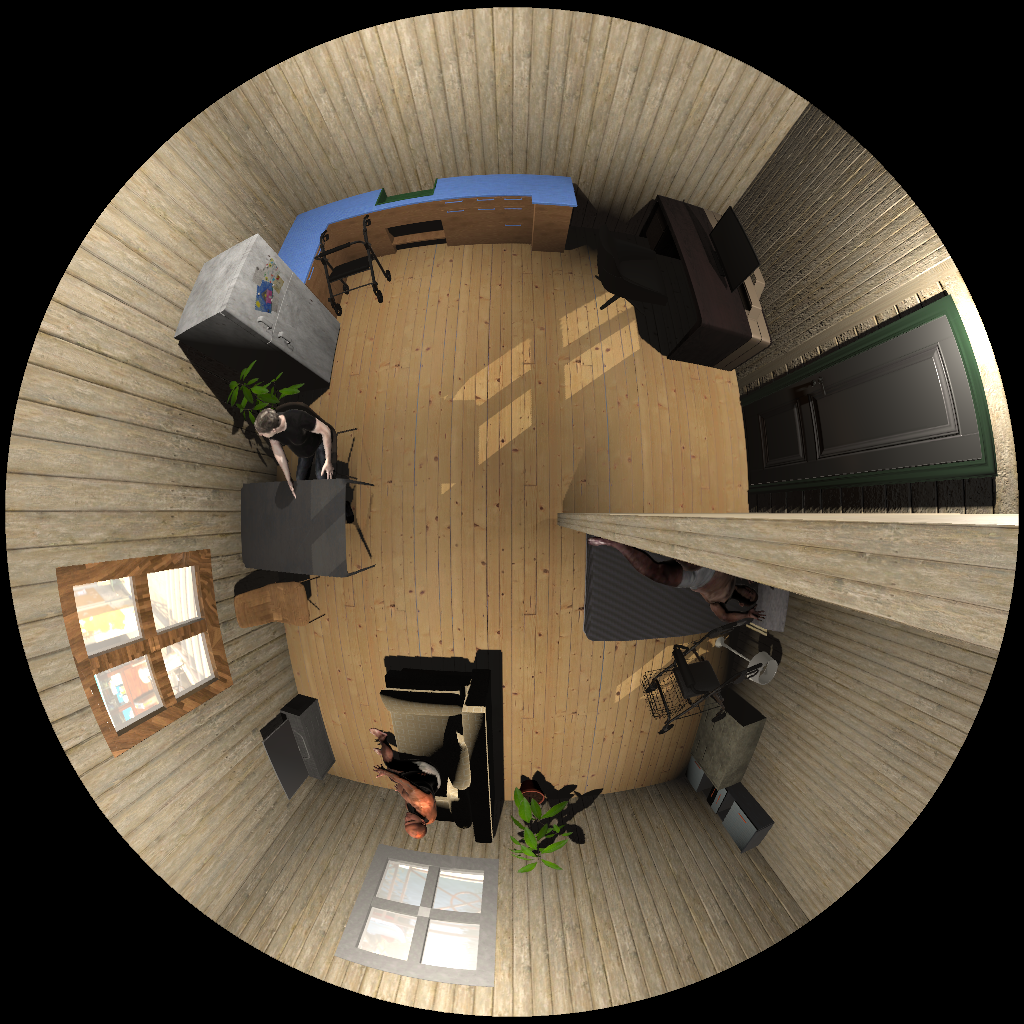}}
    %\hspace{1pt}
    \subcaptionbox{2D keypoints \label{subfig:theodore_plus:2d}}[.32\linewidth]{\includegraphics[width=\linewidth]{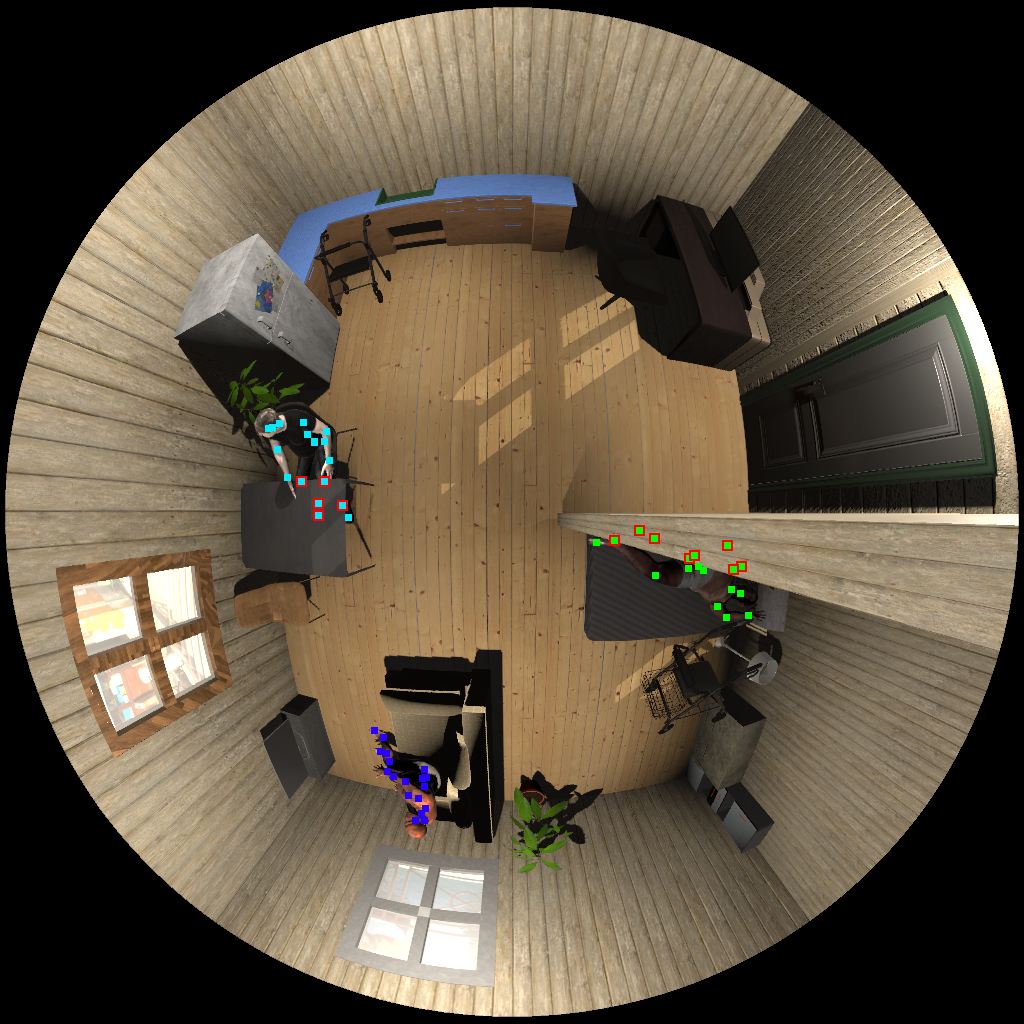}}
    %\hspace{1pt}
    \subcaptionbox{3D keypoints \label{subfig:theodore_plus:3d}}[.32\linewidth]{\includegraphics[width=\linewidth]{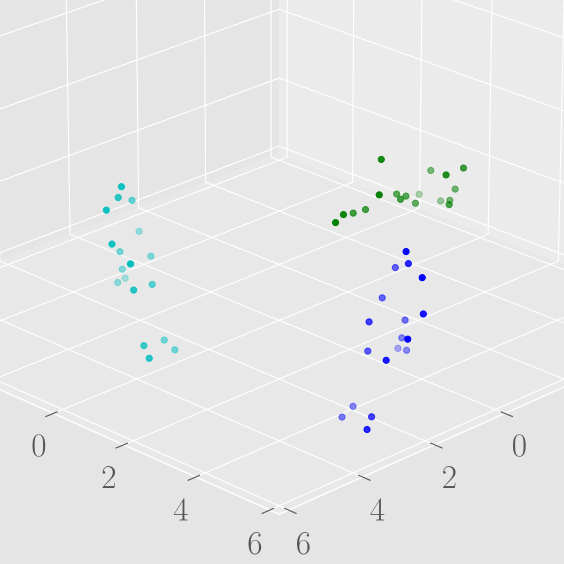}}
    \caption{Example scene from THEODORE+ dataset with an RGB image, 2D and 3D keypoints visualized. Annotations are grouped per instance (person) by colour.}
    \label{fig:theodore_plus}
\end{figure*}

Building upon the work of Scheck \etal \cite{scheck2020learning}, a large-scale synthetic dataset of indoor omnidirectional scenes is generated.
The implementation details closely follow the original work, utilizing six indoor settings and domain randomization to create a diverse dataset.
The object textures and the human model parameters (height and weight) are randomized, and at the same time the camera position changes constantly to ensure varied perspectives.
During the simulation, each character has eight pre-defined animations that belong to four categories: \textit{sitting}, \textit{lying}, \textit{falling}, and \textit{walking}.
An animation is selected from this animation set and executed depending on the character's action.
For instance, when a character interacts with a chair, the sitting animation is triggered.
Furthermore, a fall may be activated on the way to the selected chair in the virtual environment, which then executes the respective animation.
The corresponding action of a virtual character is also part of the exported data and is usable for activity recognition.
Our simulation uses Unity's universal render pipeline (UDP) for rendering the images with a distortionless virtual fisheye camera.
We did not use the high-definition render pipeline (HDRP) because of restrictions of the texture package we used, and it is debatable whether using resource intensive photorealistic synthetic images for training CNNs is advantageous \cite{tremblay2018realitygap}.

Beside bounding boxes and segmentation masks, our dataset features full body 2D and 3D pose information and action information of the human model.
The implementation based on the Unity engine is extended to map the human body skeleton points from the engine's internal format to the COCO format \cite{lin2014microsoft} with 13 keypoints.
No data is available for the simulated characters for the keypoints \textit{left\_eye}, \textit{right\_eye}, \textit{left\_ear} and \textit{right\_ear}.
Therefore, these points receive the coordinates [0,0] during the export.
The new dataset consists of \num[group-separator={,}]{50000} images with a resolution of $2048\times2048$ pixel and \(\sim\)\,\num[group-separator={,}]{160000} character instances.
Each file is saved in PNG format to exclude compression artifacts in the exported images.
The 2D keypoints are converted to pixel coordinate space during export, while the 3D keypoints remain unchanged.
An overview of the dataset modalities is shown in \cref{fig:theodore_plus}.
The dataset contains a scene's RGB image (\ref{subfig:theodore_plus:rgb}), 2D Keypoints in pixel space (\ref{subfig:theodore_plus:2d}), and non-normalized 3D keypoints (\ref{subfig:theodore_plus:3d}) of each person.
Furthermore, the keypoints' occluded attribute is true if superimposed by another object and not visible to the camera (keypoints with a red border in \ref{subfig:theodore_plus:2d}).

\subsection{PoseFES dataset}
\label{subsec:fes}

\begin{figure*}
    \centering
    \subcaptionbox{scenario 1\label{fig:fesa}}[.32\linewidth]{\includegraphics[width=\linewidth]{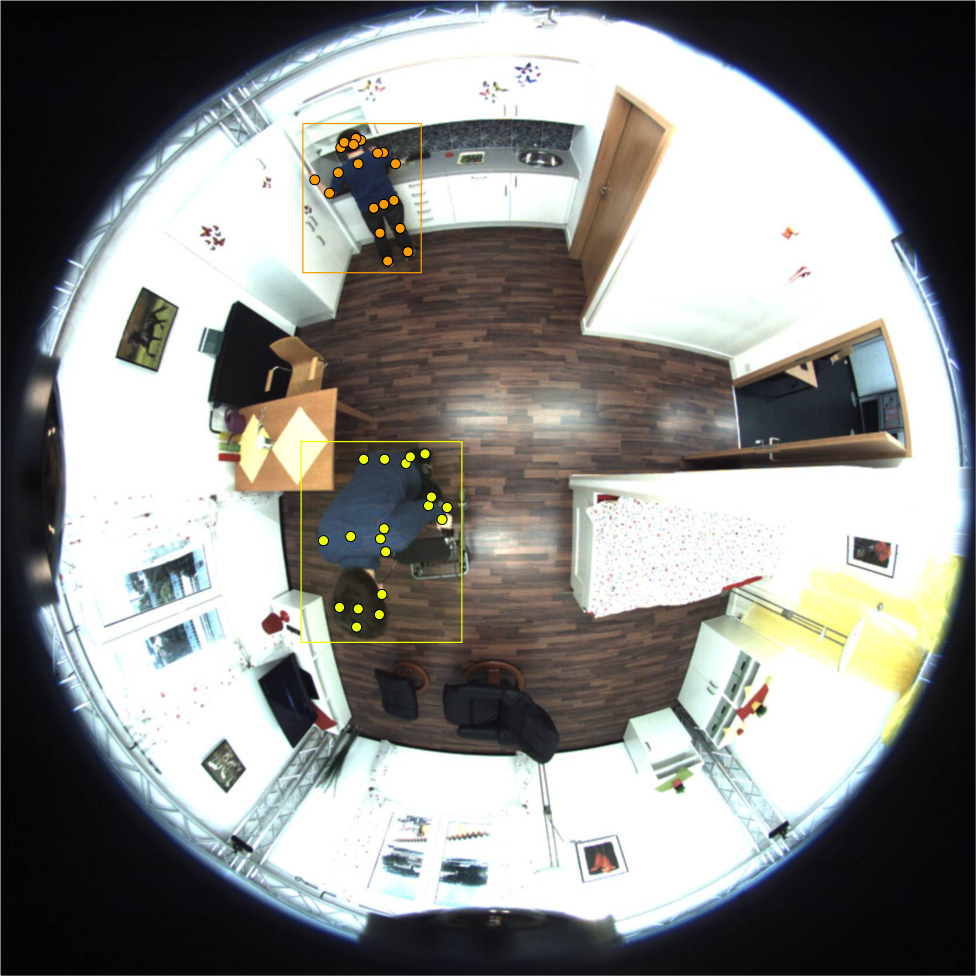}}
    \hspace{2pt}
    \subcaptionbox{scenario 2\label{fig:fesb}}[.32\linewidth]{\includegraphics[width=\linewidth]{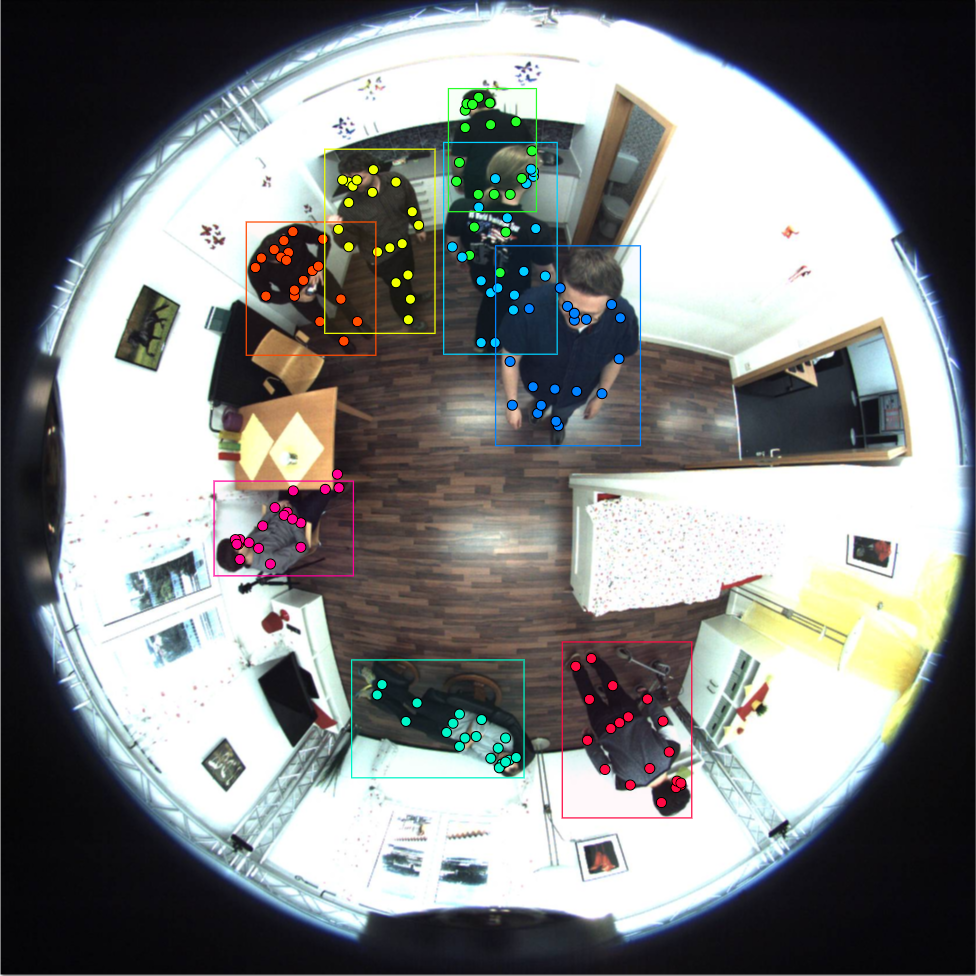}}
    \caption{Example images with annotations from PoseFES dataset. Annotations are grouped per instance (person) by colour.}
    \label{fig:labapartexample}
\end{figure*}

%The test dataset FES, which is used in \cite{scheck2020learning}, is extended with pose annotation. % not suitable for review, need anonymizing
We created the real-world dataset PoseFES for evaluation purposes by extending the FES Dataset \cite{scheck2020learning} with one scenario and pose annotations. % temporary name for review, maybe FES_Pose
It consists of two sequences, which have been recorded in a laboratory apartment with an omnidirectional fisheye camera.
The image resolution is \(1680\times1680\) pixels.
The first sequence, \textit{Scenario 1} (Sc1), contains 400 frames (Record\_00000.png -- Record\_00399.png), in which three persons walk through the apartment performing daily activities.
Overlapping of persons is kept very seldom for this sequence.
\textit{Scenario 2} (Sc2) contains 301 frames (Record\_00600.png -- Record\_00900.png), in which a maximum of eight persons appear at the same time.
Heavy overlapping is present in most frames of this sequence.
There are 735 and \num[group-separator={,}]{2161} instances in Sc1 and Sc2, respectively.

Axis-aligned bounding boxes and keypoints are annotated for the dataset.
The bounding boxes are generated using OmniPD \cite{yu2019omnipd} and then adjusted manually.
17 keypoints are annotated for all persons, which conform to the keypoints provided by COCO \cite{lin2014microsoft}.
Two extra keypoints, \textit{shoulder\_center} and \textit{hip\_center} are extrapolated by averaging shoulder and hip keypoints, respectively.
Annotations are available in CVAT and COCO format.
Sample images as well as annotations are shown in \cref{fig:labapartexample}.
More images are available in \cref{subsec:qualeval} where we present some qualitative evaluation.

\section{Training CNNs for pose estimation in omnidirectional images}
\label{sec:training}

\subsection{Model choices}
\label{subsec:chosenmodels}
We choose three models to train on our synthetic dataset.

\textbf{SimpleBaseline2D} \cite{xiao2018simple} adds three de-convolution layers with batch normalization and ReLU activation to the commonly used backbone network ResNet \cite{He_2016_CVPR}.
A \(1\times1\) convolutional layer at the end generates \(k\) heatmaps for \(k\) keypoints of a person object.
The loss is calculated by the Mean Squared Error (MSE) between the predicted heatmaps and the groundtruth heatmaps.
Using \(256\times192\) pixels as input resolution and ResNet-50 backbone, the authors are able to reach state-of-the-art performance on the COCO Keypoints dataset with this simple structure.

\textbf{HRNet} \cite{sun2019deep} introduces a new network structure for HPE.
It contains four parallel multi-resolution subnetworks.
The first stem net reduces the input resolution by 4.
The following subnetworks reduce the resolution to the half of the former one and simultaneously double the depth of the feature maps.
Information is exchanged between each stage of the subnetworks by exchange units.
At the end of the network, the heatmaps are predicted from the stem net which has the highest resolution.

Both SimpleBaseline2D and HRNet are top-down methods.
We use \textbf{MMPose} \cite{mmpose2020} to train both networks.
MMPose is a pose estimation toolbox under the OpenMMLab project.
It supports many state-of-the-art methods as well as popular datasets.
This enables us to conveniently deal with dataset manipulation and model evaluation.
We use the COCO-pretrained models provided by MMPose as our baseline, namely \textit{res50\_coco\_256x192-ec54d7f3\_20200709.pth} for SimpleBaseline2D and \textit{hrnet\_w48\_coco\_384x288-314c8528\_20200708.pth} for HRNet, both of which can be found in MMPose model zoo.

\textbf{CenterNet} \cite{zhou2019objects} employs a completely different approach for object detection and pose estimation.
Object detection is performed by estimating the heatmaps of the center points of objects and regressing the object size in \(x\) and \(y\) directions.
Keypoints are regarded as properties of the center point, and thus regressed as offset values to the center point.
In this way, CenterNet is able to perform object (in this case, person) detection and keypoint estimation for multiple objects at the same time.
It is preferable for our AAL application because of its low calculation cost, which enables it to be implemented in an embedded platform.
We use the original implementation by the authors\footnote{\url{https://github.com/xingyizhou/CenterNet}}.
The baseline models are \textit{multi\_pose\_hg\_1x.pth} and \textit{multi\_pose\_dla\_1x.pth} from its model zoo.

\subsection{Training}
The training paradigm is the same for MMPose and CenterNet:
\begin{enumerate}[label=\alph*)]
    \item The afore mentioned pretrained models are finetuned on the THEODORE+ dataset. For MMPose, learning rate (LR) is reduced to \(1/10\) of the original LR for training on COCO. For CenterNet, LR is slightly reduced from \(1.25\mathrm{e}{-4}\) to \(1\mathrm{e}{-4}\). No LR decay is used for all models.\label{train:a}
    \item All networks are trained from scratch on THEODORE+. For MMPose, LR starts at \(5\mathrm{e}{-4}\) and is decayed until \(5\mathrm{e}{-5}\). For CenterNet, the constant LR of \(7\mathrm{e}{-4}\) and \(5\mathrm{e}{-4}\) are used for the network with hourglass and DLA backbones respectively.\label{train:b}
    \item All networks are trained from scratch on the combined dataset from COCO keypoints dataset and THEODORE+ (C\&T+). LR settings are kept the same as in \ref{train:b}.\label{train:c}
    \item Pretrained models are finetuned on the combined dataset C\&T+. Settings are kept the same as in \ref{train:a}.\label{train:d}
\end{enumerate}

The pretrained models are first tested on the PoseFES dataset and the baseline performance is noted.
During finetuning or training, the models are validated directly on PoseFES after every epoch.
If the model outperforms the one from the last epoch, it is saved as the best performing model.
We finetune for 30 epochs and train for 60 epochs and choose the best performing models to fully test on PoseFES.

\section{Evaluation results on PoseFES}

\subsection{Evaluation metrics}
\label{subsec:metrics}
We use the object keypoint similarity (OKS)-based average precision (AP) and average recall (AR) as defined in COCO \cite{lin2014microsoft} to evaluate the trained models.
COCO API evaluates 17 keypoints by default.
However, THEODORE+ dataset only has 13 keypoints.
Due to catastrophic forgetting \cite{french1999catastrophic}, the model is unable to estimate the positions of eyes and ears after training or finetuning solely with THEODORE+ dataset.
Therefore, we evaluate all the models on the 13 keypoints, excluding eyes and ears.
In this case, the COCO API is adapted by deleting the sigma values for eyes and ears.
Sigma values for other keypoints are kept unchanged.
The models that are trained or finetuned with the combined dataset (training routines \ref{train:b} and \ref{train:c}) are evaluated additionally on all 17 keypoints.
Unfortunately, there exists no commonly used large scale dataset for evaluating HPE in top-down view of a fisheye camera.
The PanopTOP31K dataset \cite{garau2021panoptop} is not suitable due to the  shortcomings mentioned in \cref{subsec:tvdatasets} and the fact that the images are not generated with a fisheye camera model.
Therefore, the performance of all models are evaluated based on their performance on PoseFES dataset.

For SimpleBaseline2D and HRNet, the person bounding boxes are required for evaluation, since they are top-down methods.
We provide the models with ground truth bounding boxes, which means the person detection accuracy is 100\si{\percent}.
To estimate the influence of the person detection accuracy, we tested the models finetuned on C\&T+ using bounding boxes that are inferenced by OmniPD\cite{yu2019omnipd}, whose accuracy on PoseFES is 85.6\si{\percent}.
CenterNet performs person detection and keypoints estimation at the same time, therefore it does not need bounding box input.
%Because of this, the performance of SimpleBaseline2D and HRNet are not directly comparable to CenterNet.
%Therefore, we look into the evaluation results of MMPose models and CenterNet models separately.

\subsection{Evaluation results of MMPose models}
\label{subsec:evalMM}

\begin{figure}[h]
    \centering
    \includegraphics[width=\linewidth]{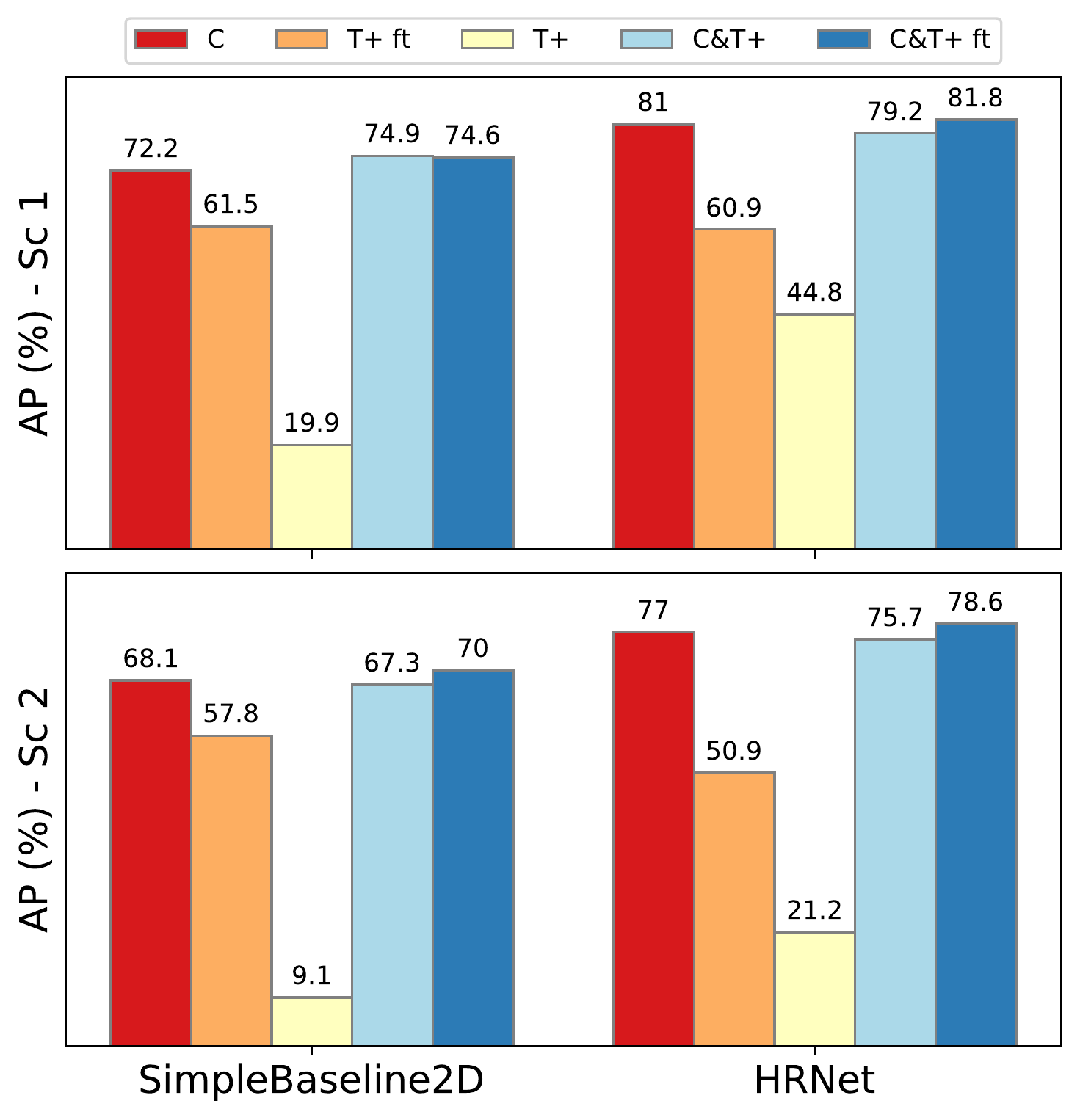}
    \caption{Comparison of MMPose models before and after training, evaluated on PoseFES for 13 KPs. Top diagram shows evaluation AP on Sc1 and bottom diagram shows evaluation AP on Sc2. \textit{C} corresponds to COCO-pretrained models, \textit{T+ ft}, \textit{T+}, \textit{C\&T+} and \textit{C\&T+ ft} correspond to training routines \ref{train:a} - \ref{train:d}, respectively. \textit{ft} stands for finetuning. AP is given at \(\textrm{OKS}=0.5:0.05:0.95\).}
    \label{fig:13kp_mmpose}
\end{figure}

We first evaluate the COCO pretrained models and our trained models on the PoseFES Dataset for 13 keypoints on scenario 1 and 2 separately.
\cref{fig:13kp_mmpose} visualizes the evaluation APs of MMPose models on PoseFES as bar charts, so that the trends can be easily spotted. Commonly for MMPose models \textit{SimpleBaseline2D} and \textit{HRNet}, finetuning pretrained models on THEODORE+ alone results in heavy performance degradation.
For both scenarios, AP is reduced by 10\si{\percent} for SimpleBaseline2D and over 20\si{\percent} for HRNet.
Predictably, training from scratch on THEODORE+ alone yields very low performance.
Finetuning on the combined dataset results in performance improvements of \SIrange[range-phrase = --]{1}{2}{\percent} for both models.
The two models behave differently when trained from scratch on the combined dataset.
For HRNet, the trained model is slightly worse than the COCO pretrained model, while for SimpleBaseline2D the results are slightly better than the finetuned model for scenario 1 but worse for scenario 2.
The complete evaluation results are shown in \cref{tab:eval13kpsMM}.
The evaluation results using OmniPD bounding boxes are denoted C\&T+ ft w/PD.

\begin{table}[h]
    \centering
    \caption{Evaluation results of MMPose models for 13 KPs on PoseFES Dataset. The best results are marked bold.
        %HRNet yields best performance when the pretrained model is finetuned with the combined dataset.
        %SimpleBaseline2D is less consistant for Sc1 and 2.
        }
        {\small
        \begin{tabular}{c|l|C{1cm}C{1cm}|cc}
            & Model & \multicolumn{2}{c|}{SimpleBaseline2D} & \multicolumn{2}{c}{HRNet}\\
            \midrule
            Sc & Dataset  & AP & AR & AP & AR \\
            \midrule
            \multirow{6}{*}{1} & COCO & 72.2 & 74.8 & 81.0 & 83.0 \\
                            & T+ ft & 61.5 & 63.6 & 60.9 & 63.1 \\
                            & T+ & 19.9 & 23.6 & 44.8 & 47.6 \\
                            & C\&T+ & \textbf{74.9} & 77.6 & 79.2 & 81.6 \\
                            & C\&T+ ft & 74.6 & \textbf{76.8} & \textbf{81.8} & \textbf{83.7} \\
                            & C\&T+ ft w/ PD & 60.1 & 61.5 & 65.4 & 66.7 \\
            \midrule
            \multirow{6}{*}{2} & COCO & 68.1 & \textbf{74.8} & 77.0 & 79.2 \\
                            & T+ ft & 57.8 & 61.5 & 50.9 & 62.2 \\
                            & T+ & 9.1 & 13.8 & 21.2 & 27.8 \\
                            & C\&T+ & 67.3 & 70.6 & 75.7 & 77.7 \\
                            & C\&T+ ft & \textbf{70.0} & 72.6 & \textbf{78.6} & \textbf{80.7} \\
                            & C\&T+ ft w/ PD & 59.0 & 60.5 & 63.8 & 64.9 \\
            \bottomrule
        \end{tabular}
        }
    \label{tab:eval13kpsMM}
\end{table}

Scenario 2 is more difficult for both models.
Overall, the highest AP and AR values are reached by finetuning pretrained models on the combined dataset.
For HRNet, the improvement is 1.6\si{\percent} for AP and 1.5\si{\percent} for AR.
It is less consistent for SimpleBaseline2D.

\begin{table}[h]
    \centering
    \caption{Evaluation results of MMPose models for 17 KPs on PoseFES Dataset. The best results are marked bold.}
        {\small
        \begin{tabular}{c|l|C{1cm}C{1cm}|cc}
            & Model & \multicolumn{2}{c|}{SimpleBaseline2D} & \multicolumn{2}{c}{HRNet} \\
            \midrule
            Sc & Training  & AP & AR & AP & AR \\
            \midrule
            \multirow{4}{*}{1} & COCO & \textbf{67.4} & \textbf{69.9} & 75.7 & 78.2 \\
                            & C\&T+ & 66.9 & 69.7 & 72.2 & 75.0 \\
                            & C\&T+ ft & \textbf{67.4} & 69.7 & \textbf{76.1} & \textbf{78.3} \\
                            & C\&T+ ft w/ PD & 53.7 & 55.3 & 60.4 & 62.1 \\
            \midrule
            \multirow{4}{*}{2} & COCO & \textbf{63.2} & \textbf{67.0} & 71.9 & 74.6 \\
                            & C\&T+ & 60.0 & 63.4 & 68.4 & 71.7 \\
                            & C\&T+ ft & 62.6 & 65.8 & \textbf{72.9} & \textbf{75.3} \\
                            & C\&T+ ft w/ PD & 52.3 & 54.3 & 58.9 & 60.2 \\
            \bottomrule
        \end{tabular}
        }
    \label{tab:eval17kpsMM}
\end{table}

\cref{tab:eval17kpsMM} lists evaluation results for 17 keypoints.
For HRNet the results stay in line with those for 13 keypoints.
Finetuning pretrained model on combined dataset yields the best performance, albeit less than 1\si{\percent} over COCO pretrained model.
For SimpleBaseline2D, training or finetuning on the combined dataset provides no benefit over COCO pretrained model.

\subsection{Evaluation results of CenterNet models}
\label{subsec:evalCN}

\begin{figure}[h]
    \centering
    \includegraphics[width=\linewidth]{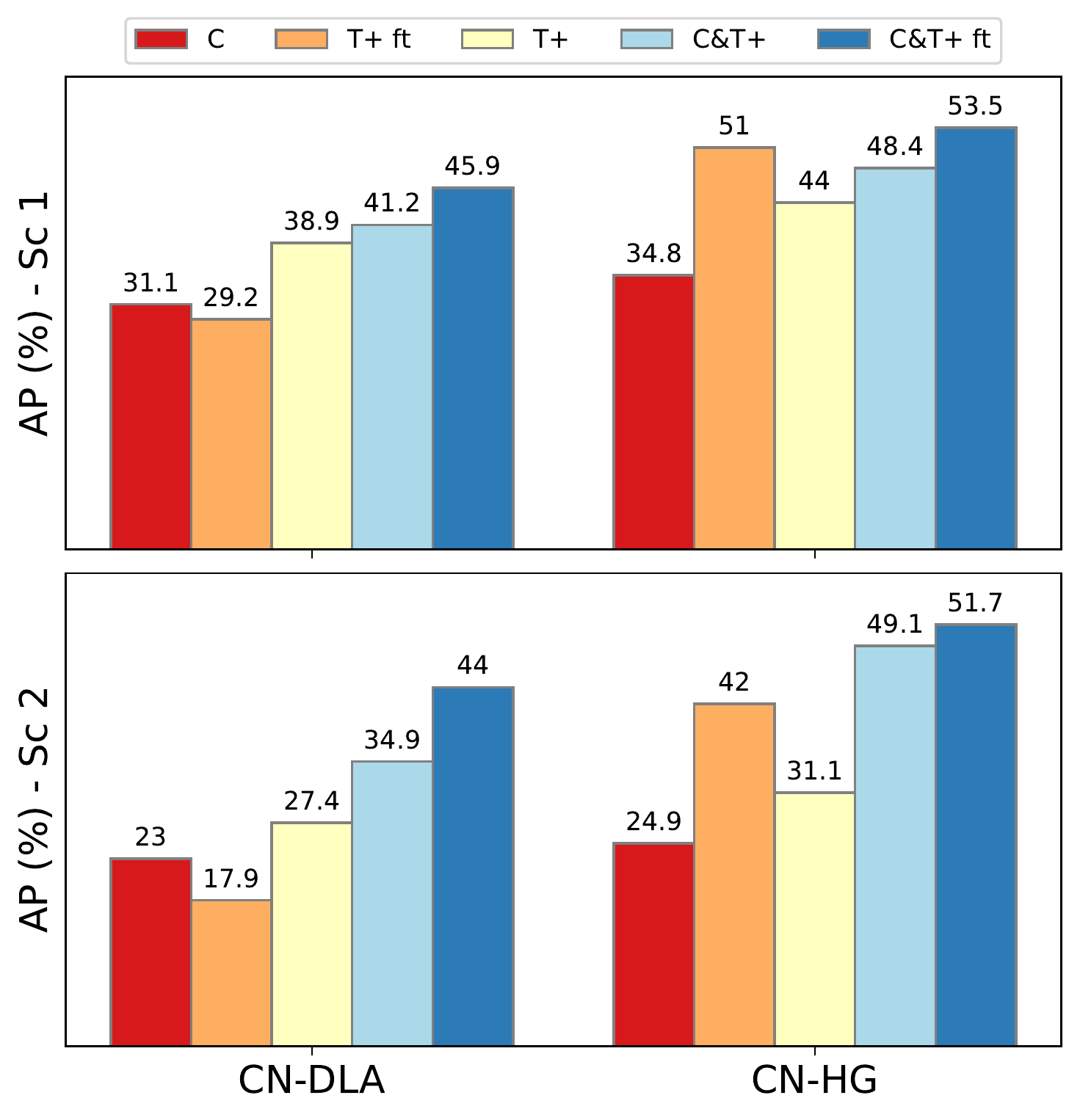}
    \caption{Comparison of CenterNet models before and after training, evaluated on PoseFES for 13 KPs. \textit{CN-DLA} stands for CenterNet with DLA\cite{yu2018deep} backbone, and \textit{CN-HG} stands for CenterNet with Hourglass\cite{newell2016stacked} backbone.}
    \label{fig:13kp_centernet}
\end{figure}

\begin{table}[h]
    \centering
    \caption{Evaluation results of CenterNet models for \textbf{13 KPs} on PoseFES Dataset. The best results are marked bold.}
        {\small
        \begin{tabular}{c|l|cc|cc}
            & Model & \multicolumn{2}{c|}{CN-DLA} & \multicolumn{2}{c}{CN-HG} \\
            \midrule
            Sc & Dataset & AP & AR & AP & AR \\
            \midrule
            \multirow{5}{*}{1} & COCO & 31.1 & 35.0 & 34.8 & 38.4 \\
                            & T+ ft & 29.2 & 34.8 & 51.0 & 55.5\\
                            & T+ & 38.9 & 44.4 & 44.0 & 48.8 \\
                            & C\&T+ & 41.2 & 44.4 & 48.4 & 52.6 \\
                            & C\&T+ ft & \textbf{45.9} & \textbf{50.7} & \textbf{53.5} & \textbf{56.6}\\
            \midrule
            \multirow{5}{*}{2} & COCO & 23.0 & 27.6 & 24.9 & 30.4 \\
                            & T+ ft & 10.4 & 17.9 & 34.3 & 42.0 \\
                            & T+ & 17.6 & 27.4 & 21.5 & 31.1 \\
                            & C\&T+ & 26.7 & 34.9 & 39.5 & 49.1 \\
                            & C\&T+ ft & \textbf{34.5} & \textbf{44.0} & \textbf{43.9} & \textbf{51.7}\\
            \bottomrule
        \end{tabular}
        }
    \label{tab:eval13kpsCN}
\end{table}

\begin{table}[h]
    \centering
    \caption{Evaluation results of CenterNet models for \textbf{person detection} on PoseFES Dataset.}
        {\small
        \begin{tabular}{c|l|cc|cc}
            & Model & \multicolumn{2}{c|}{CN-DLA} & \multicolumn{2}{c}{CN-HG} \\
            \midrule
            Sc & Dataset & AP & AR & AP & AR \\
            \midrule
            \multirow{2}{*}{1} & COCO & 44.2 & 52.7 & 47.9 & 55.2 \\
%                            & T+ ft & 47.0 & 54.1 & 51.1 & 56.6\\
%                            & T+ & 47.3 & 52.8 & 50.1 & 56.0 \\
%                            & C\&T+ & \textbf{49.6} & \textbf{57.2} & 53.1 & 58.5 \\
                            & C\&T+ ft & 48.4 & 56.0 & 53.8 & 62.0\\
            \midrule
            \multirow{2}{*}{2} & COCO & 56.9 & 64.3 & 54.5 & 61.9 \\
%                            & T+ ft & 45.4 & 52.8 & 60.1 & 66.9 \\
%                            & T+ & 53.2 & 58.8 & 53.0 & 60.7 \\
%                            & C\&T+ & 60.2 & 65.6 & 68.0 & 73.7 \\
                            & C\&T+ ft & 65.9 & 71.7 & 69.8 & 74.4\\
            \bottomrule
        \end{tabular}
        }
    \label{tab:evalbboxCN}
\end{table}

CenterNet shows very different behaviours than MMPose models.
In \cref{fig:13kp_centernet}, both variants benefit from joint training of object detection and keypoints estimation when evaluated on PoseFES.
Both models significantly outperform the COCO pretrained models except for CN-DLA when finetuned only on THEODORE+.
To isolate the influence of the improvement in person detection on HPE, we compare the evaluation results of keypoint estimation in \cref{tab:eval13kpsCN} to the evaluation results of bounding boxes in \cref{tab:evalbboxCN}.
To keep it simple, we compare only the models of finetuning with the combined dataset with the COCO pretrained model.
CN-DLA's improvement in person detection AP is 4.2\si{\percent} for Sc1 and 9\si{\percent} for Sc2. 
Its improvement in keypoint estimation AP is 14.8\si{\percent} for Sc1 and 11.5\si{\percent} for Sc2.
For CN-HG, the person detection AP improvement lies at 5.9\si{\percent} for Sc1 and 15.3\si{\percent} for Sc2.
Its improvement in keypoint estimation AP is even more significant at 18.7\si{\percent} for Sc1 and 19\si{\percent}.
We can conclude that the relationship between improvement in person detection and keypoint estimation is not proportional.
Keypoint estimation benefits much more from the training than person detection.

Evaluation results for 17 keypoints are listed in \cref{tab:eval17kpsCN}.
Similarly to 13 keypoints, both models perform best when COCO pretrained model is finetuned on combined COCO and THEODORE+ dataset.

\begin{table}[h]
    \centering
    \caption{Evaluation results of CenterNet models for 17 KPs on PoseFES Dataset. The best results are marked bold.}
        {\small
        \begin{tabular}{c|l|cc|cc}
            & Model & \multicolumn{2}{c|}{CN-DLA} & \multicolumn{2}{c}{CN-HG} \\
            \midrule
            Sc & Training & AP & AR & AP & AR \\
            \midrule
            \multirow{3}{*}{1} & COCO & 28.1 & 31.7 & 32.4 & 35.7 \\
                            & C\&T+ & 29.6 & 32.7 & 34.4 & 38.4 \\
                            & C\&T+ ft & \textbf{35.1} & \textbf{40.3} & \textbf{38.1} & \textbf{42.4}\\
            \midrule
            \multirow{3}{*}{2} & COCO & 19.9 & 24.7 & 21.7 & 27.0 \\
                            & C\&T+ & 15.1 & 22.4 & 24.1 & 34.4 \\
                            & C\&T+ ft & \textbf{23.3} & \textbf{33.3} & \textbf{27.7} & \textbf{36.1}\\
            \bottomrule
        \end{tabular}
        }
    \label{tab:eval17kpsCN}
\end{table}

\subsection{Qualitative evaluation}
\label{subsec:qualeval}

In this section we evaluate the quality of the estimation results by the two most relevant models, HRNet and CenterNet with Hourglass backbone, both finetuned on C\&T+.

\begin{figure}
    \includegraphics[width=\linewidth]{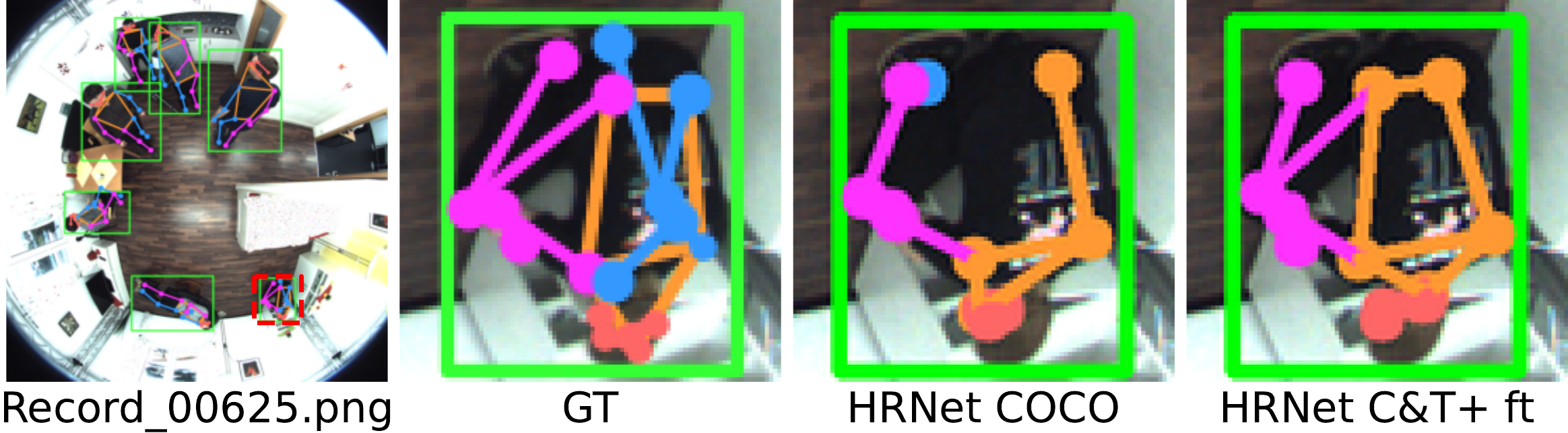}
    \includegraphics[width=\linewidth]{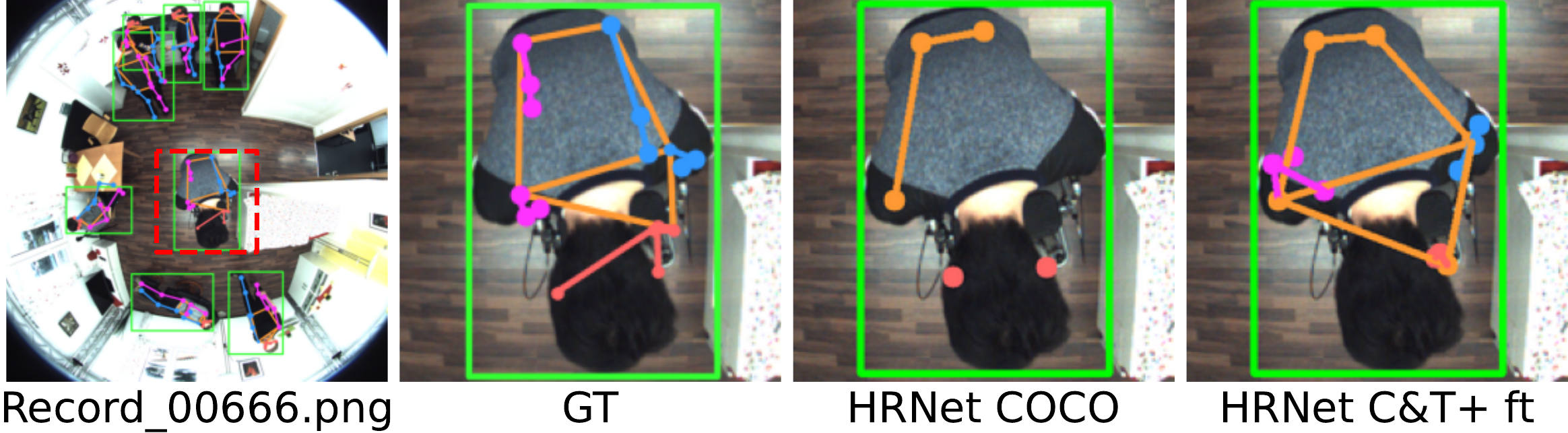}
    \includegraphics[width=\linewidth]{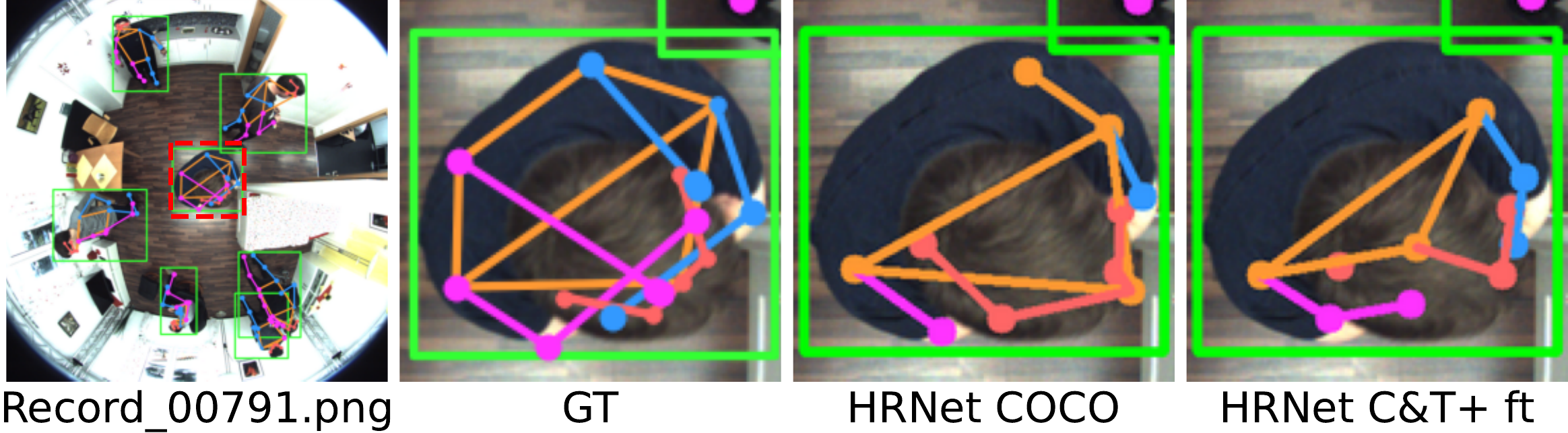}
    \caption{Comparison of estimation examples by \textbf{HRNet} finetuned on C\&T+. The cropped area are marked with red dashed bounding boxes in the original image.}
    \label{fig:qualitativehrnet}
\end{figure}

We conclude from the quantitative evaluation that the finetuning provides only an improvement of the AP by less than 2\si{\percent} for \textbf{HRNet},
which seems not significant.
Persons at the edges or in the outer circle of the omnidirectional view look similar to those in the perspective image when standing or lying down.
These instances are therefore not difficult for a good top-down model like HRNet even without finetuning.
However, the appearance of the body changes dramatically when the person bends over or goes to the center of the omnidirectional image as shown in \cref{fig:qualitativehrnet}.
Our finetuned model shows a clear improvement over the original model in these critical cases.
In addition, we observe improvements in estimating joints in occluded body parts either by other body parts of the same person or by another person.

\begin{figure}[b]
    \centering
    \includegraphics[width=\linewidth]{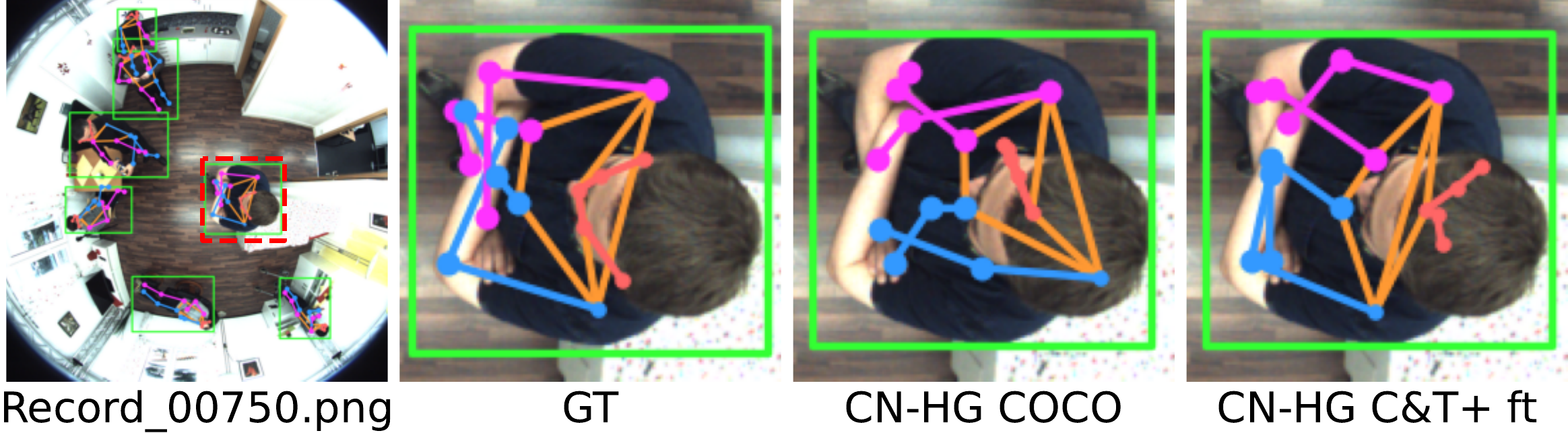}
    \includegraphics[width=\linewidth]{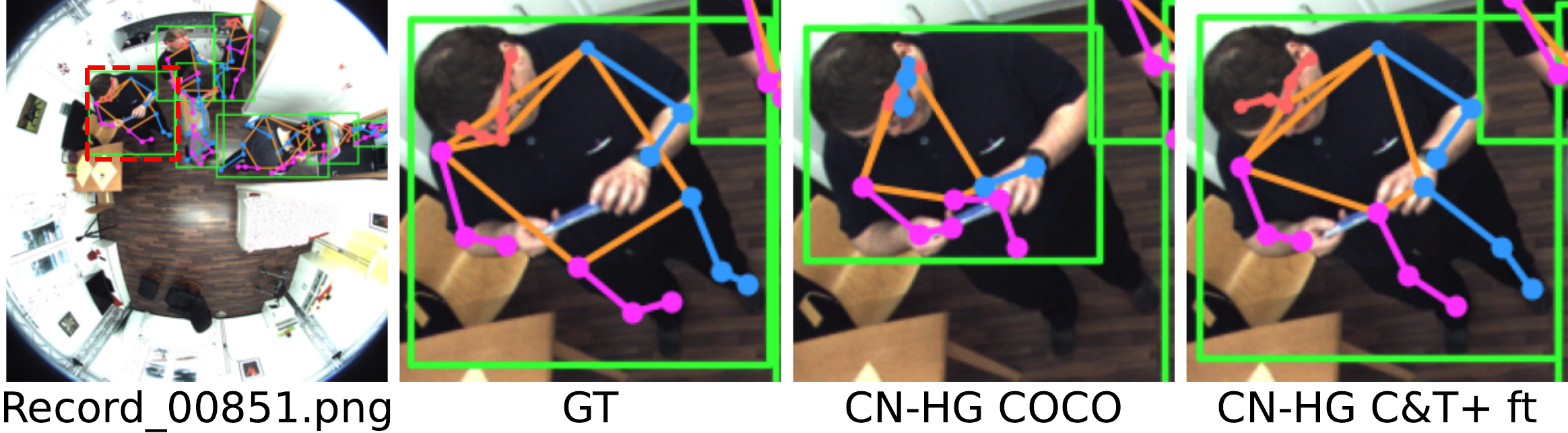}
    \includegraphics[width=\linewidth]{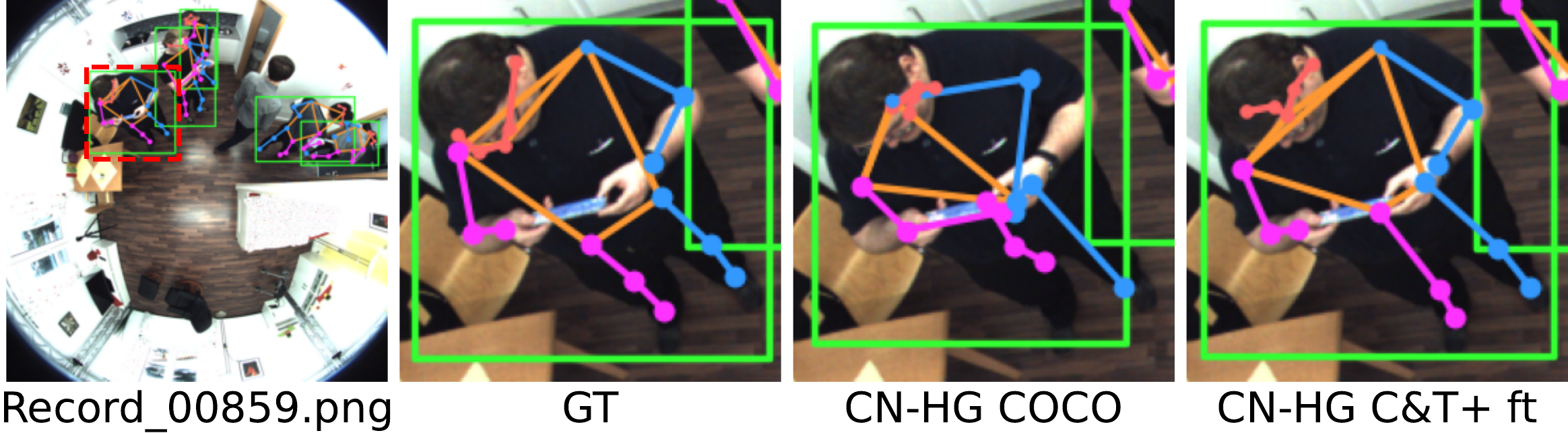}
    \caption{Instances of improved estimation results by CenterNet.}
    \label{fig:qualitativecenternet}
\end{figure}

\begin{figure}
    \centering
    \includegraphics[width=\linewidth]{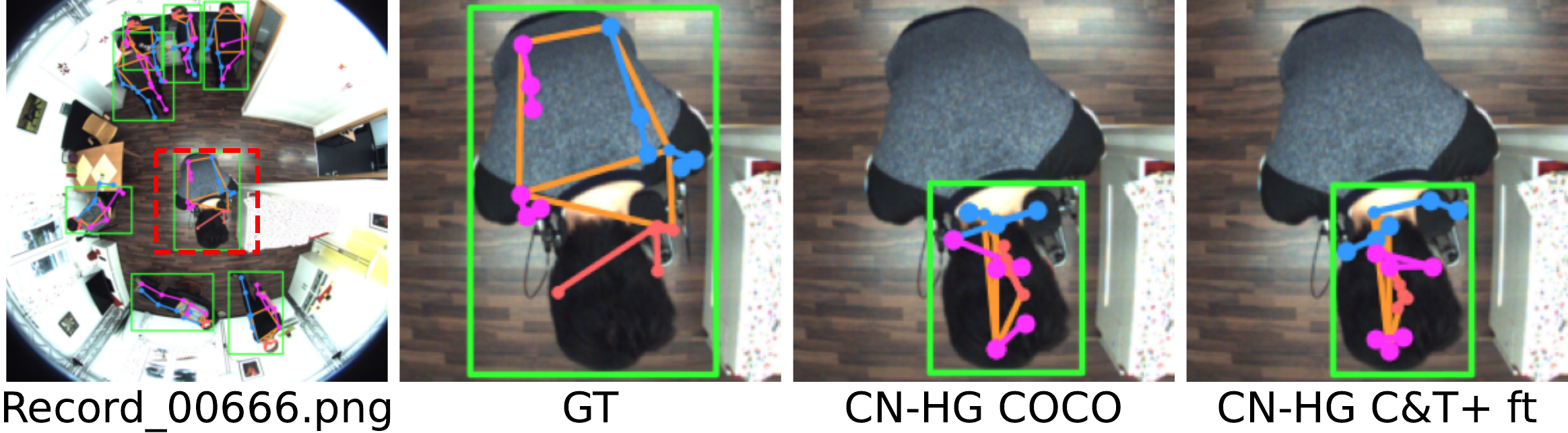}
    \caption{A failed detection and joint estimation by CenterNet.}
    \label{fig:qualfailcenternet}
\end{figure}

The improvement is more evident for the \textbf{CenterNet} model with Hourglass backbone. 
At the confidence threshold of 0.4, the finetuned model can reliably detect persons in the scene while keeping false positives at minimum.
Smaller instances around the edges and instances near the center of the image are detected more reliably than the COCO pretrained model.
The estimated keypoints are more accurate, especially for instances in the center directly under the camera.
\cref{fig:qualitativecenternet} shows better estimation results by the finetuned CenterNet model because of more accurate person detection and more precise joint estimation.
However, the CenterNet model clearly falls behind the HRNet model.
Estimation of occluded joints is hardly improved over COCO pretrained model.
Failure cases such as in \cref{fig:qualfailcenternet} is often the result of faulty person detection.
Undetected person naturally means no joint estimation, for example the instance in Rechord\_00625.png shown in \cref{fig:qualitativehrnet} cannot be detected by the finetuned CenterNet model.

\subsection{Temporal performance}
Our test bench is a workstation with an Intel Core i9-9960X CPU and 128\,GB of DDR4 memory.
The GPU used for inference is an Nvidia Titan RTX with 24\,GB of GDDR memory.

The MMPose models uses groundtruth bounding boxes, therefore the inference time for person detection is not taken into account.
HRNet at the resolution of \(384\times288\) pixels needs on average \SI{214}{\milli\second} per image for inferencing on Sc1 of PoseFES and \SI{574}{\milli\second} on Sc2, which makes an average of \SI{369}{\milli\second} per image on the whole dataset.
The inference time depends on the number of instances, as is the same with other top-down methods.
The average inference time per instance is \SI{89}{\milli\second} for HRNet.
The corresponding values for SimpleBaseline2D is \SI{159}{\milli\second} per image for Sc1, \SI{468}{\milli\second} per image for Sc2, \SI{291}{\milli\second} per image for the whole dataset and \SI{71}{\milli\second} per instance at the resolution of \(256\times192\) pixels.
Using OmniPD for person detection adds about \(20\)\,\(\sim\)\,\(30\)\,\si{\milli\second} overhead per image to the top-down pipeline.

CenterNet is a bottom-up model and its speed does not depend on the number of instances per image.
The measured inference time for CenterNet with DLA backbone is \SI{53}{\milli\second} per image, and for CenterNet with Hourglass backbone it is \SI{173}{\milli\second}.
Both models process the inputs at the resolution of \(512\times512\) pixels.

\section{Conclusion}
\label{sec:conclusion}
In this paper, we provide THEODORE+, a new synthetic omnidirectional top-view dataset with \num[group-separator={,}]{50000} RGB-images and annotated 2D and 3D keypoints, containing four action categories: \textit{sitting}, \textit{lying}, \textit{falling} and \textit{walking}, rendered through the unity engine.
Furthermore, for evaluation purposes, the PoseFES dataset, which contains 701 images of real-world omnidirectional indoor scenes was manually annotated with 2D keypoints.
We then trained and finetuned state-of-the-art top-down and bottom-up HPE models for perspective images to successfully perform HPE on PoseFES dataset, namely SimpleBaseline2D, HRNet and CenterNet with two different backbone networks.

% The evaluation of THEODORE+ was done on 13 and 17 keypoints through state-of-the-art CNNs for keypoint detection, namely MMPose and CenterNet.
% Our baseline is on COCO pretrained models.
% Finetuning on the combined dataset from COCO keypoints and THEODORE+ delivers the best results on 13 keypoints for all models.
% SimpleBaseline2D and HRNet in MMPose increase in AP by 1.9\si{\percent} and 1.6\si{\percent} for the harder Sc2 of PoseFES.
% Qualitative evaluation reveals that the improvement is most evident for instances of the top-view.
% Both variants of CenterNet get improved by over 11\si{\percent}, while the variant with Hourglass backbone get the highest improvement of AP and AR by over 18.7\si{\percent} for both scenarios with respect to the baseline.
% They benefit from joint training of object detection and keypoint estimation.
% The evaluation on 17 keypoints shows similar results for HRNet and CenterNet, yet with lower absolute improvement.

The key findings of our work can be summarized as follows. 
We figured out that (1) the training on synthetic data from a \emph{perfect} omnidirectional camera model significantly improve the results of real-world data, captured by a \emph{non-calibrated} omnidirectional camera.
Therefore, the network is able to deal with the distortion from different omnidirectional camera models.
(2) The recognition of persons and keypoints close to the center of the image are much more improved than side-viewed persons. 
This corresponds to the top-view task, which is one of our motivation to generate THEODORE+ for the home monitoring of elderly.
Our key finding (3) is already observed in the work of Scheck \etal \cite{scheck2020learning}, and it could be confirmed through our work that synthetic object detection bounding boxes improve the results for both disciplines, top-view object detection and top-view keypoint estimation.

Optimizing our best trained model to an embedded platform for person monitoring in an AAL-context is our major next step.
Further investigations based on THEODORE+ will be activity recognition and the lifting of 2D human poses to 3D poses with and without stereo image data.
Thanks to our finetuned models, it is possible to create a large-scale real-world human pose estimation dataset for omnidirectional top-view with limited resources in the future.
Our datasets and the synthetic data generation pipeline will be open for download after the publication of this paper.

{\small
   \bibliographystyle{ieee_fullname}
   \bibliography{yjinlit}
}

\end{document}